\pdfoutput=1

\documentclass{article}
\usepackage{ijcai24}
\usepackage{chemfig}
\usepackage{times}
\usepackage{soul}
\usepackage{url}
\usepackage[hidelinks]{hyperref}
\usepackage[utf8]{inputenc}
\usepackage[small]{caption}
\usepackage{graphicx}
\usepackage{amsmath}
\usepackage{amsthm}
\usepackage{booktabs}
\usepackage{algorithm}
\usepackage{algorithmic}
\usepackage[switch]{lineno}
\usepackage{colortbl}
\usepackage{color}
\usepackage{multirow}
\usepackage{enumitem}
\usepackage{graphicx}
\usepackage{svg}
\usepackage{amssymb}
\usepackage{subfiles}

\title{Spatio-Temporal Field Neural Networks for Air Quality Inference}

\author{
Yutong Feng$^1$\and Qiongyan Wang$^1$\and Yutong Xia$^2$ \and Junlin Huang$^1$\and Siru Zhong$^1$
\and Yuxuan Liang$^{1,3}$\footnote{Corresponding author. Email: yuxliang@outlook.com}
\affiliations
$^1$Hong Kong University of Science and Technology (Guangzhou), China\\
$^2$National University of Singapore, Singapore\\
$^3$State Key Lab of Resources and Environmental Information System, China\\
\emails
{\{yfeng083,jhuang688,szhong691\}}@connect.hkust-gz.edu.cn,\\
\{{yutong.x,qiongyanwang,yuxliang\}}@outlook.com
}

\begin{document}

\maketitle

\begin{abstract}
    Air quality inference aims to utilize historical data from a limited number of observation sites to infer the air quality index at unknown locations. Considering data sparsity due to the high maintenance cost of stations, good inference algorithms can effectively save the cost and refine the data granularity. While spatio-temporal graph neural networks have made excellent progress on this problem, their non-Euclidean and discrete data structure modeling of reality limits its potential. In this work, we make the first attempt to combine two different spatio-temporal perspectives, fields and graphs, by proposing a new model, Spatio-Temporal Field Neural Network, and its corresponding new framework, Pyramidal Inference. Extensive experiments validate that our model achieves state-of-the-art performance in nationwide air quality inference in the Chinese Mainland, demonstrating the superiority of our proposed model and framework. 
\end{abstract}

\section{Introduction}
Real-time monitoring of air quality, such as PM2.5, PM10, and \chemfig{NO_2} concentrations, is crucial for air pollution control and protecting human health, with air pollution contributing to seven million deaths annually according to the WHO~\cite{vallero2014fundamentals}. However, the deployment of air quality stations in urban areas is limited due to high costs, requiring around 200,000 USD for construction and 30,000 USD annually for maintenance~\cite{zheng2013u}. Additionally, these stations need significant land and dedicated personnel for upkeep, further limiting their prevalence in cities.
\begin{figure}[t]
    \centering
    \includegraphics[width=\linewidth]{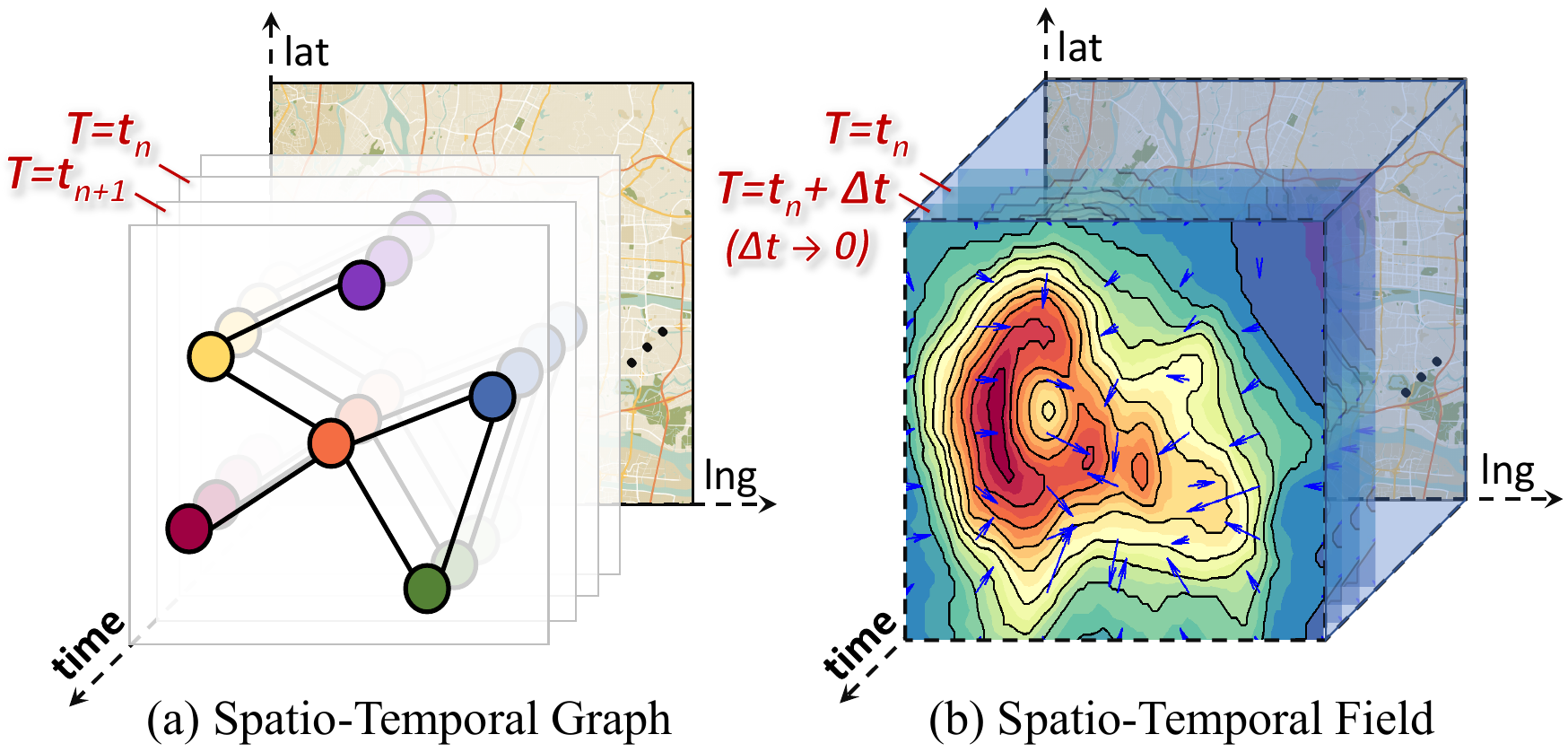}
    \caption{Spatio-Temporal Graph vs. Spatio-Temporal Field.}
    \label{fig:enter-label}
\end{figure}
\begin{figure*}[!t]
  \centering
  \includegraphics[width=0.9\linewidth]{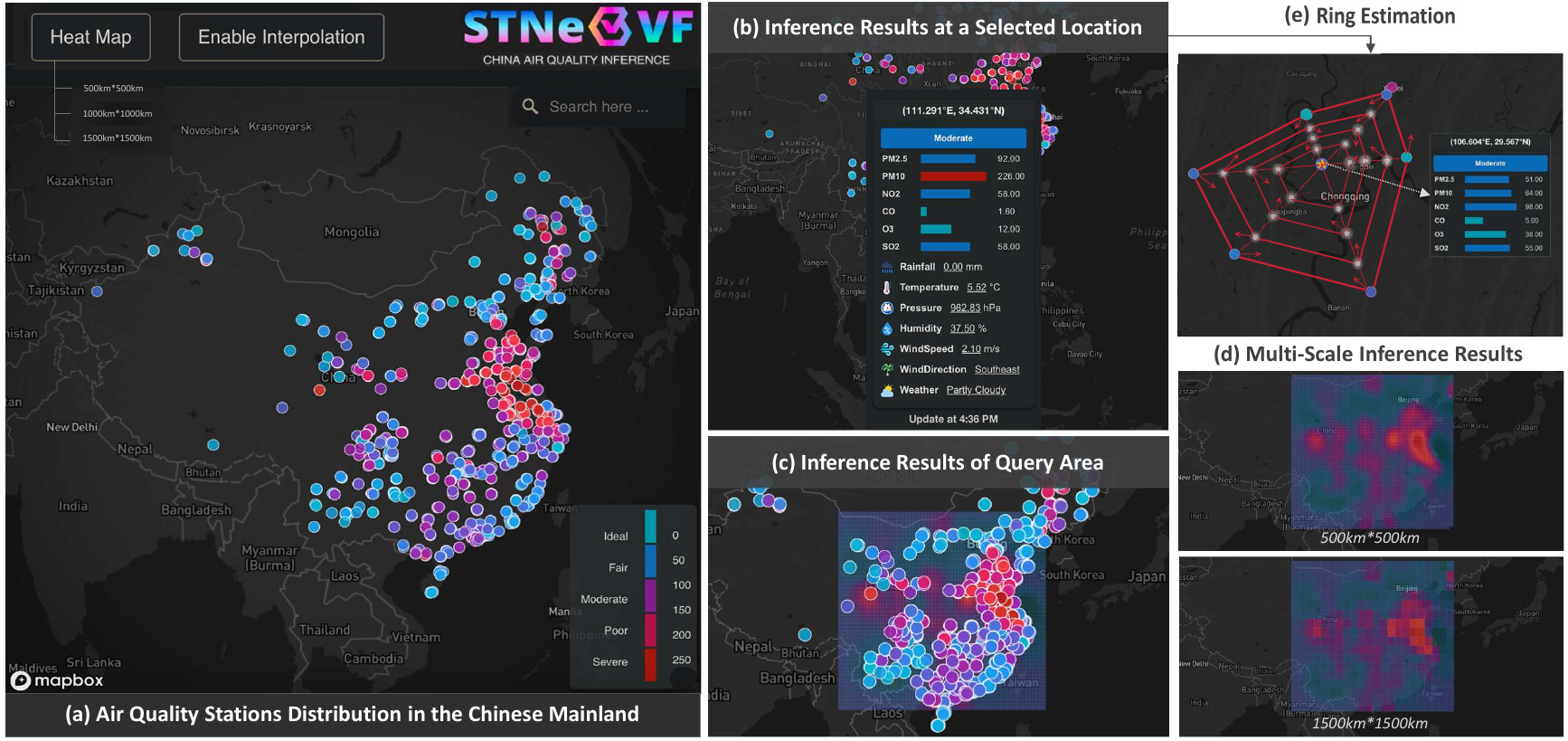}
  \vspace{-0.3em}
  \caption{(a)-(d): User interface of our STFNN system for air quality inference. (e): An illustration of Ring Estimation.}
  \label{fig:intro}
  \vspace{-1em}
\end{figure*}


In the past decade, substantial research endeavors have been directed towards \emph{air quality inference}~\cite{han2023machine}, seeking to infer real-time air quality in locations devoid of monitoring stations by leveraging data gleaned from existing sites, as shown in Figure \ref{fig:intro}(b)-(c). 
With recent advancements in deep learning, Graph Neural Networks (GNN)~\cite{kipf2016semi} have become dominant for non-Euclidean data representation, particularly in learning complex spatial correlations among air quality monitoring stations. Integrating GNNs with temporal learning modules (e.g., RNN~\cite{graves2013generating}, TCN~\cite{bai2018empirical}, ODE~\cite{liang2022mixed}) has led to the development of \emph{Spatio-Temporal Graph Neural Networks} (STGNN)\cite{wang2020deep,jin2023spatio}, addressing the dynamic nature of air quality data across spatial and temporal dimension. STGNNs, exemplified in studies like~\cite{han2021fine,hu2023decoupling} offer superior representation extraction and flexibility in cross-domain data fusion.

Though promising, STGNNs simply treat air quality data as a Spatio-Temporal Graph (STG), as shown in Figure \ref{fig:enter-label}(a). However, these models overlook a crucial property – \emph{continuity}, which manifests across both spatial and temporal dimensions. In reality, air quality readings of stations are sampled from a continuous Euclidean space and cannot be fully encapsulated by a discrete graph structure using GNNs. Meanwhile, the temporal modules (e.g., RNN, TCN) in STGNNs exhibit the discrete nature as well, rendering them incapable of capturing continuous-time dynamics within data. To better represent the continuous and evolving nature of real-world air quality phenomena, a more powerful approach is needed, surpassing the discrete representation of STGNNs.

In this paper, we draw inspiration from Field Theory~\cite{mcmullin2002origins} and innovatively formulate air quality inference from a \emph{field} perspective, where air quality data is a physical quantity that can be conceptualized by a new concept called \emph{Spatio-Temporal Fields} (STF), as depicted in Figure \ref{fig:enter-label}(b). These fields encompass three dimensions (i.e., latitude, longitude, time), assigning a distinct value to each point in spacetime. In contrast to STGs, STFs are characterized by being \textbf{regular}, \textbf{continuous}, and \textbf{unified}\footnote{It implies that the field representation accounts for variations not only across different locations in space but also over different points in time, emphasizing the comprehensive treatment of both spatial and temporal aspects within the unified framework.}, offering a representation more aligned with reality. Under this perspective, we can transform the air quality inference problem to \emph{reconstruct} STFs from available readings using coordinate-based neural networks, particularly Implicit Neural Representations (INR)~\cite{sitzmann2020implicit,xie2022neural}.


While INRs effectively handle the continuity property of air quality data, they inevitably confront two primary challenges. Firstly, the generation process of air quality data is extremely complex and influenced by various factors (such as humidity and wind speed/direction), which poses a challenge for reconstructing the underlying STFs through neural representation methods. Secondly, empirical studies~\cite{xu2019frequency,sitzmann2020implicit} verify that INRs always exhibit a bias towards learning low-frequency functions, which will disregard locally varying high-frequency information and higher-order derivatives even with dense supervision.

To this end, we for the first time present \textbf{Spatio-Temporal Field Neural Networks} (\textbf{STFNN}), opening new avenues for modeling spatio-temporal fields and achieving state-of-the-art performance in nationwide air quality inference in the Chinese Mainland. Targeting the first challenge, we pivot our focus from reconstructing the value of each entry in STFs to learning the derivative (i.e., gradient) of each entry. This strategic shift is inspired by learning the residual is often easier than learning the original value directly, as exemplified in ResNet~\cite{he2016deep}. Such vector field can not only show how the pollutant concentration varies across time and space but also the direction of diffusion. To tackle the second challenge, we endeavor to augment our STFNN with local context knowledge during air quality inference at a specific location. Specifically, we combine the STGNN's capability to capture local spatio-temporal dependencies with STFNN's ability to learn global spaito-temporal unified representations. This integration results in what we term \textbf{Pyramid Inference}, a hybrid framework that leverages the strengths of both models to achieve a more comprehensive inference of air quality dynamics with both high-frequency and low-frequency components. Overall, our contributions lie in three aspects:
\begin{itemize}[leftmargin=*]
\item \emph{A Field Perspective}. We formulate air quality as spatio-temporal fields with the first shot. Compared to STGs, our STFs not only adeptly capture the continuity and Euclidean structure of air quality data, but also achieves a unified representation across both space and time.

\item \emph{Spatio-Temporal Field Neural Networks}. We propose a groundbreaking network called STFNN to model STF data. STFNN pioneers an implicit representation of the STF's gradient,  deviating from conventional direct estimation approaches. Moreover, it preserves high-frequency information via Pyramid Inference.

\item\emph{Empirical Evidence}. We conduct extensive experiments to evaluate the effectiveness of our STFNN. The results valiadte that STFNN outperforms prior arts by a significant margin and exhibits compelling properties. A system in Figure \ref{fig:intro} has been deployed to show its practicality in the Chinese Mainland.

\end{itemize}

\section{Preliminary}
\noindent\textbf{Definition 1 (Air Quality Reading)} We use $\mathbf{x}_t^i \in \mathbb{R}^D$ and $\mathbf{y}_t^i \in \mathbb{R}$ to denote the air quality readings and the concentration of PM2.5 from the $i$-th monitoring stations at time $t$ separately. Here $D$ encompasses various measurements, such as concentrations of other air pollutants (e.g., PM10, \chemfig{NO_2}), and meteorological properties (e.g. humidity, weather and wind speed). $\mathbf{X}_t = \left( \mathbf{x}_t^1,\mathbf{x}_t^2,\hdots,\mathbf{x}_t^N \right) \in \mathbb{R}^{N \times D}$ denotes the observations of all stations at a specified time $t$. $\mathcal{X} = \left( \mathbf{X}_1,\mathbf{X}_2,\hdots,\mathbf{X}_T \right) \in \mathbb{R}^{T \times N \times D}$ denotes the observations of all stations at all time. Similar definitions apply to $\mathbf{Y}_t$ and $\mathcal{Y}$, mirroring $\mathbf{X}_t$ and $\mathcal{X}$, respectively.

\noindent\textbf{Definition 2 (Coordinates)} A coordinate $\mathbf{c}=\left[lng,lat,t\right] \in \mathbb{R}^3$ is used to represent the spatial and temporal properties of an air quality reading or a location, including longitude, latitude, and timestamp. These coordinates are categorized into two types: \textit{source} coordinate $\mathbf{c}^{src}$, associated with readings or locations with existing air quality monitoring stations, while \textit{target} coordinate $\mathbf{c}^{tar}$, corresponding to unobserved locations requiring inference. Notably, $\mathbf{c}_t^i$ represents the coordinate of the corresponding $\mathbf{x}_t^i$ and $\mathbf{y}_t^i$. Parallel definitions apply to $\mathbf{C}_t$ and $\mathcal{C}$ in relation to $\mathbf{X}_t$ and $\mathcal{X}$, respectively, which are not reiterated here.

\begin{figure}[h]
    \centering
    \includegraphics[width=\linewidth]{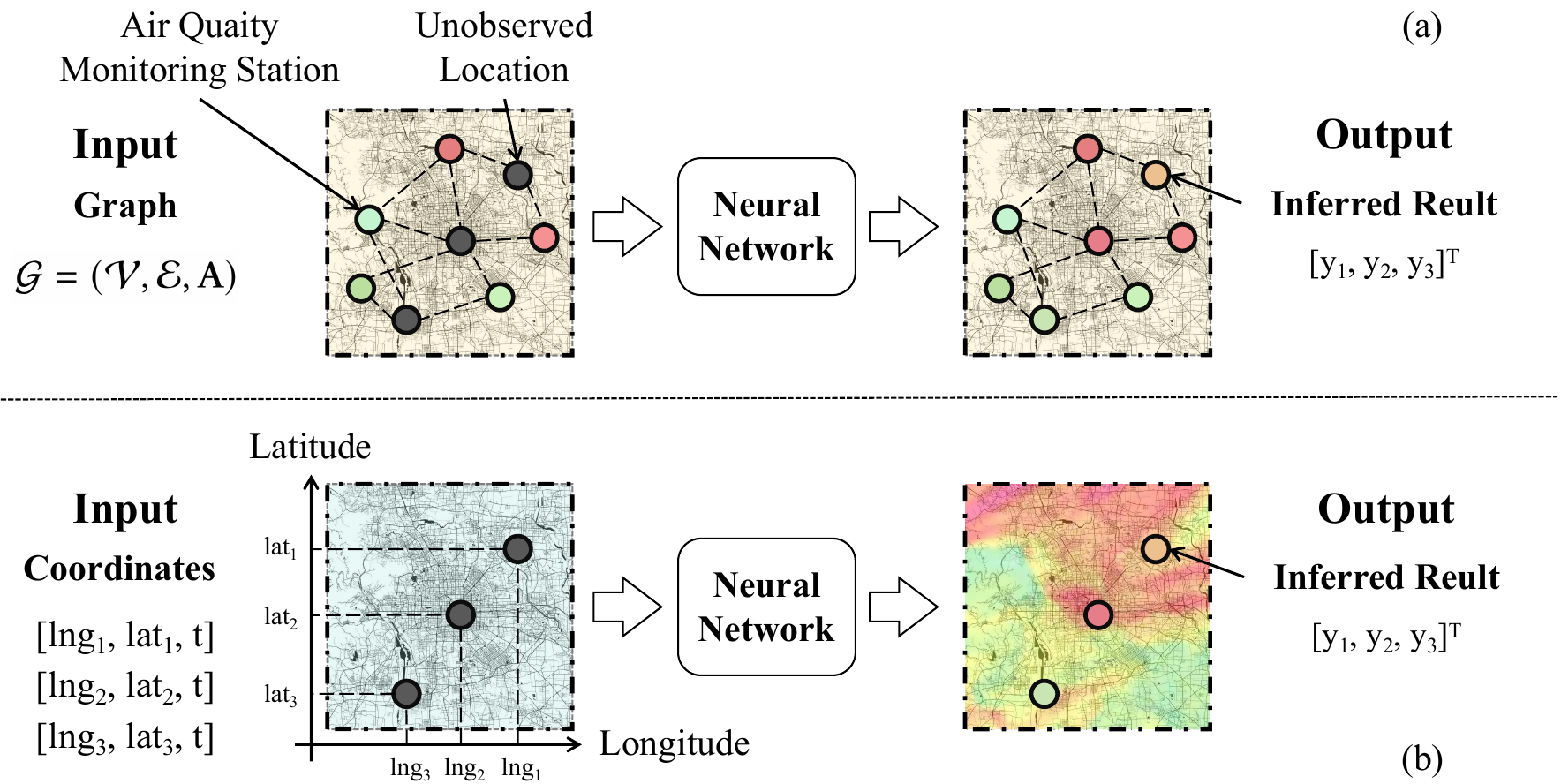}
    \caption{Paradigms for air quality inference. (a) A spatio-temporal graph perspective. (b) A spatio-temporal field perspective.}
    \label{problem definition}
\end{figure}

\noindent\textbf{Problem Definition} The air quality inference problem addresses the utilization of historical data and real-time readings from a limited number of air quality monitoring stations to infer the real-time air quality \emph{anywhere}, especially unobserved location. 
Traditional strategies~\cite{hou2022graphmae,hu2023graph} employ graphs to illustrate the relationship between stations and locations, and the task is translated into a recovery task for the masked nodes (target locations), as shown in Figure \ref{problem definition} (a).
In this paper, we revisit the problem from the field perspective, as shown in Figure \ref{problem definition} (b). Specifically, our goal is to reconstruct a spatio-temporal field $G$ for air quality that is capable of mapping any arbitrary coordinate, especially $\mathbf{c}^{tar}$, to the corresponding concentration of PM2.5 $\mathbf{y}^{tar}$. Additional parameters, such as $\mathcal{X}$, are allowed to enhance the inference process.

\section{Methodology}
\subsection{Global View: Spatio-Temporal Field}
A Spatio-Temporal Field (STF) is a global modeling of air quality that encompasses all stations and observation times. The STF function, denoted as $f(\cdot):\mathbf{c} \longmapsto \mathbf{q}$, assigns a unique physical quantity $\mathbf{q}$ to each coordinate. When $\mathbf{q}$ is a scalar, $f(\cdot)$ represents a scalar field. Conversely, if $\mathbf{q}$ is a vector, with magnitude and direction, it denotes a vector field. 

Specifically, our focus lies on a scalar field $G: \mathbb{R}^3 \rightarrow \mathbb{R}$ for air quality inference, where $G$ maps the coordinates to the corresponding PM2.5 concentration. This representation facilitates a continuous and unified spacetime perspective, allowing for the inference of air quality at any location and time by inputting the coordinates.
 
Directly modeling $G$ is challenging due to its intricate complexity and nonlinearity. Alternatively, it is often more feasible to learn its derivative, which refers to the gradient field of $G$ in spacetime. We denote the gradient field as $\mathbf{F} \triangleq \nabla G$ which is a vector field. Notably, given a specific $\mathbf{F}$, an array of $G$ solutions exists unless an initial value is specified. We use $\mathbf{y}^{src}$ and $\mathbf{y}^{tar}$ to denote the PM2.5 concentration on $\mathbf{c}^{src}$ and $\mathbf{c}^{tar}$, respectively. Our primary focus lies in inferring $\mathbf{y}^{tar}$ since the true value of $\mathbf{y}^{src}$ is known and recorded while $\mathbf{y}^{tar}$ remains undisclosed. To infer $\mathbf{y}^{tar}$, we utilize a $\mathbf{y}^{src}$ as the initial value and assume $l$ is a piecewise smooth curve in $\mathbb{R}^3$ that point from $\mathbf{c}^{src}$ to $\mathbf{c}^{tar}$, then we have
\begin{equation}\label{curve integration}
\begin{split}
\mathbf{y}^{tar} &= G\left( \mathbf{c}^{tar} \right) = G\left( \mathbf{c}^{src} \right) + \int_{l}\nabla G(\mathbf{r})\cdot d\mathbf{r} \\
&= \mathbf{y}^{src} + \int_{a}^{b}\mathbf{F}\big(\mathbf{r}(z)\big)\cdot\mathbf{r}^{\prime}(z)dz 
\end{split}
\end{equation}
where $\cdot$ is the dot product, and $\mathbf{r}: \left[a, b\right] \rightarrow l$ represents the position vector. The endpoints of $l$ are given by $\mathbf{r}(a)$ and $\mathbf{r}(b)$, with $a < b$. Seeking an Implicit Neural Representation (INR) becomes the objective to fit $\mathbf{F}$ as $\mathbf{F}$ is usually intricate to the extent that it cannot be explicitly formulated. 



\subsection{Local View: Spatio-Temporal Graph}
The formulation presented in Eq. (\ref{curve integration}) ensures the recoverability of $\mathbf{c}^{tar}$ across arbitrary coordinates through curve integration, utilizing solely a single initial value. However, this approach yields an excessively coarse representation of the entire STF, resulting in the loss of numerous local details and high-frequency components
The impact is particularly pronounced when $\mathbf{y}^{src}$ is situated at a considerable distance from $\mathbf{y}^{tar}$ as the increase in the length of $l$ introduces a significant cumulative error. 
In response to this limitation, we leverage the potent learning capabilities of STGNN to capture local spatio-temporal correlations effectively. We employ the local spatio-temporal graph (STG) to model the spatio-temporal dependencies of the given coordinates and their neighboring air monitoring stations with their histories. The corresponding design can be found in the foundational work \cite{song2020spatial}, and it is not necessary to reiterate it here.

\begin{figure*}[!t]
  \centering
  \includegraphics[width=\linewidth]{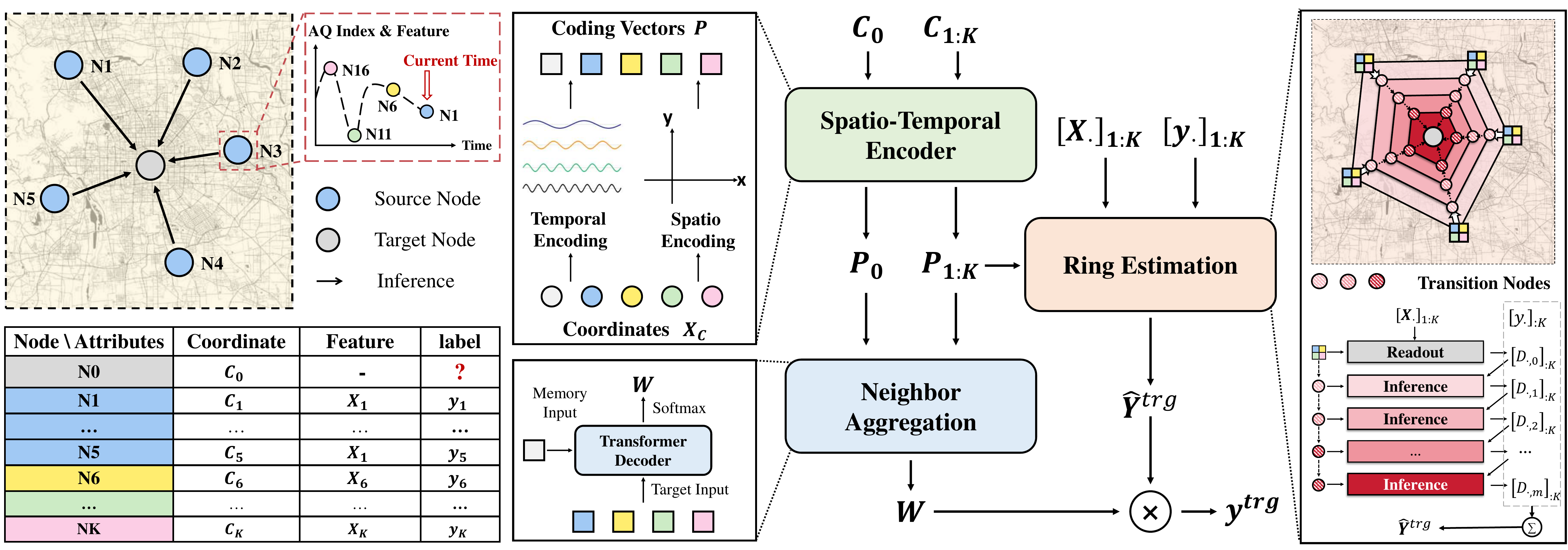}
  \caption{Implementation of STFNN}
  \label{model struct}
\end{figure*}
\subsection{Hybrid Framework: Pyramidal Inference}
We intend to integrate the continuous and uniform global modeling of spacetime provided by STF with the local detailing capabilities of STG, thereby establishing a hybrid framework that leverages the strengths of both approaches. Within the local STG, the estimation of $\mathbf{y}^{tar}$ is achieved by leveraging information from its neighboring nodes through Eq. (\ref{curve integration}). By calculating estimates of $\mathbf{y}^{tar}$ from these neighbors and assigning a learnable weight $w_i$ to each estimation ($\sum_{i=1}^{K}w_i=1$), we enhance the precision of the inference result for a specific coordinate. This operation can be formulated as
\begin{equation}\label{pi eq}
\begin{split}
    \hat{\mathbf{y}}^{tar} &= \sum_{i=1}^{K} \left[w_i \cdot \left( \mathbf{y}_i^{src} + \int_{l_i}\mathbf{F}(\mathbf{r})\cdot d\mathbf{r} \right) \right]\\
    &= \sum_{i=1}^{K} \left[w_i \cdot \left( \mathbf{y}_i^{src} + \int_{a_i}^{b_i}\mathbf{F}\big(\mathbf{r}(z)\big)\cdot\mathbf{r}^{\prime}(z)dz \right) \right]
\end{split}
\end{equation}
where $\hat{\mathbf{y}}^{tar}$ is the joint estimation of $\mathbf{y}^{tar}$ by neighbors. $w_i$ and $\mathbf{y}_i^{src}$ represent the weight and PM2.5 concentration of the $i_{th}$ neighbor, respectively. $l_i$ is the integral path in $\mathbb{R}^3$ that points from the coordinates of the $i_{th}$ neighbor to the target coordinate. 

We call the inference strategy represented by Eq. (\ref{pi eq}) \textbf{Pyramidal Inference}, which is the framework of the STFNN we propose. To demonstrate its sophistication, we deconstruct Eq. (\ref{pi eq}) into two steps:
\begin{equation}\label{fa eq}
    \hat{\mathbf{Y}}^{tar} = \left[
    \begin{array}{c}
        \hat{\mathbf{y}}_1 \\
         \vdots \\
         \hat{\mathbf{y}}_{K}
    \end{array}
    \right]
    =\left[
    \begin{array}{c}
         \mathbf{y}_1^{src} + \int_{\mathcal{C}_i}\mathbf{F}(\mathbf{r})\cdot d\mathbf{r} \\
         \vdots \\
         \mathbf{y}_{K}^{src} + \int_{\mathcal{C}_K}\mathbf{F}(\mathbf{r})\cdot d\mathbf{r} 
    \end{array}
    \right]
\end{equation}
and
\begin{equation}\label{sta eq}
    \hat{\mathbf{y}}^{tar} = \left[
    \begin{array}{ccc}
         w_1 & \hdots & w_K
    \end{array}
    \right] \cdot \hat{\mathbf{Y}}^{tar}
    = 
    \mathbf{W}^T \cdot \hat{\mathbf{Y}}^{tar},
\end{equation}
where $\hat{\mathbf{y}}_i$ is the estimate of $\mathbf{y}^{tar}$ by the $i_{th}$ neighbor, $\mathbf{W}$ and $\hat{\mathbf{Y}}^{tar}$ represent the vector form of $w_i$ and $\hat{\mathbf{y}}_i$, respectively. These two operations can be viewed as follows. In Eq. (\ref{fa eq}), a curve integral is used over the gradient field to estimate $\mathbf{y}^{tar}$ from each neighbor in a continuous and spatio-temporally uniform way, which takes advantage of the STF. In Eq. (\ref{sta eq}), the information from the neighbors is aggregated through $\mathbf{W}$, which considers the spatio-temporal dependencies between the nodes and takes advantage of the STG. In this way, the Pyramidal Inference framework combines the two different spacetime perspectives into a distinctive new paradigm.

\section{Implementation}

Upon introducing the formulation of Pyramidal Inference, we proceed to its implementation through a meticulously designed model architecture, depicted in Figure \ref{model struct}. The model comprises three pivotal components: 
\begin{itemize}[leftmargin=*]
    \item \textbf{Spatio-Temporal Encoding.} This component transforms coordinates into coded vectors endowed with representational meaning, enhancing the network's ability to comprehend and leverage the spatio-temporal characteristics of coordinates.
    \item \textbf{Ring Estimation:} Implementation of Eq. (\ref{fa eq}), mapping the encoded vector to the gradient of the STF. This process yields each neighbor's estimate of the PM2.5 concentration for the target coordinates through a path integral.
    \item \textbf{Neighbor Aggregation:} Implementation of Eq. (\ref{sta eq}), utilizing the coded vectors of neighbors and target coordinates as inputs to derive the estimated weights for each neighbor concerning the target coordinates.
\end{itemize}
In the following parts, we will provide a detailed exposition of each module, elucidating their functionalities step by step.

\subsection{Spatio-Temporal Encoding}\label{st encoding sec}
We revisit the previously introduced local STG, which amalgamates nodes across different time steps into a unified graph, potentially obscuring the inherent temporal properties of individual nodes. In essence, this local STG places nodes from diverse time steps into a shared environment without discerning their temporal distinctions. However, this issue can be mitigated through meticulous positional coding of nodes \cite{gehring2017convolutional,song2020spatial}.

We use $\mathbf{p} \in \mathbb{R}^{10}$ to denote the coding vector, essential for accurately describing the spatio-temporal characteristics of a node or coordinate. This vector is expressed as the concatenation $\mathbf{p} = \left[\mathbf{p}_S,\mathbf{p}_T\right]$, where $\mathbf{p}_S$ represents spatial coding and $\mathbf{p}_T$ represents temporal coding. In the spatial dimension, a node's properties can be captured by its absolute position, represented by z-normalized longitude $lng_z$ and latitude $lat_z$, forming $\mathbf{p}_S = \left[lng_z,lat_z\right]$. For encoding temporal information $\mathbf{p}_T$, sinusoidal functions with different periods are employed, reflecting the periodic nature of temporal phenomena. We utilize a set of periods $\textbf{T} = \left\{ 1a,7a,30.5a,365a \right\}$, with $a$ as the scaling index, to represent days, weeks, months, and years. The temporal coding $\mathbf{p}_T$ is then represented as
\begin{equation}
    \mathbf{p}_{(T,i)} =
        \begin{cases}
        sin\left(2\pi t/\mathbf{T}_{int(i/2)+1}\right) & \text{$i \bmod 2=0$}\\
        cos\left(2\pi t/\mathbf{T}_{int(i/2)+1}\right) & \text{$i \bmod 2\neq0$}
        \end{cases}
\end{equation}
where $\mathbf{p}_{(T,i)}$ denotes the value of the $i_{th}$ dimension ($1\leq i \leq 8$) of $\mathbf{p}_T$, $int(i/2)$ denotes dividing $i$ by 2 and rounding down, and $\mathbf{T}_{int(i/2)+1}$ is the $int(i/2)+1$ period of $\mathbf{T}$.  

\subsection{Ring Estimation}\label{ring inference}
 \noindent\textbf{Motivation.}  We employ the continuous approach of curve integration over the gradient to determine the PM2.5 concentration at the target coordinate in Eq. (\ref{fa eq}). However, due to inherent limitations in numerical accuracy within computing systems, achieving true continuity in STF becomes unattainable. Therefore, we have adopted an incremental approach. We set the integral path to be a straight line from the neighbors' coordinate $\mathbf{c}^{src}$ to the target coordinate $\mathbf{c}^{tar}$ for convenience, then the unit direction vector $\vec{\mathbf{r}}$ in the path can be written as $\vec{\mathbf{r}} \triangleq \left( \mathbf{c}^{tar}-\mathbf{c}^{src} \right) \big/ \left( \Vert \mathbf{c}^{tar}-\mathbf{c}^{src} \Vert \right)$. After that, we replace the integral operation with a summation operation and modify Eq. (\ref{fa eq}) to a discrete form
\begin{equation}\label{dis pi}
\begin{split}
    \hat{\mathbf{y}}^{tar} &= \sum_{i=1}^{K} \left[w_i \cdot \left( \mathbf{y}_i^{src} + \sum_{j=1}^{m} \mathbf{D}_{i,j} \cdot \vec{\mathbf{r}_i} \right) \right]\\
\end{split}
\end{equation}
where $m$ represents the step size of the summation and $\mathbf{c}_i^{src}$ is the coordinate of the $i_{th}$ node in the local STG. $\mathbf{D}_{i,j}\in \mathbb{R}^3$ represents the \emph{difference} at the $j_{th}$ step of the $i_{th}$ node, which is the discrete approximation of the gradient. Our objective is to build a module for estimating $\mathbf{D}_{i,j}$, which is the only unknown in Eq. (\ref{dis pi}).

\noindent\textbf{Overview.}  Towards this objective, we introduce a pivotal module named Ring Estimation, designed for the joint estimation of the differences $\left[\mathbf{D}_{\cdot,j} \right]_{:K}=\left[\mathbf{D}_{1,j},\cdots,\mathbf{D}_{K,j} \right]\in \mathbb{R}^{K \times 3} $ at the $j_{th}$ step of all neighbors. We posit that simultaneous estimation of $\left[\mathbf{D}_{\cdot,j} \right]{:K}$ enhances inference efficiency and captures correlations between them, thereby reducing estimation errors compared to individually estimating $\mathbf{D}_{i,j}$ for a single neighbor $K$ times. Specifically, the Ring Estimation module divides the polygon surrounded by $K$ neighbors around the target coordinate into $m$ ring zones from the outermost to the innermost, with total $mK$ of \emph{transition} nodes (coordinates) uniformly spaced along the inference path. The inner edge of the $j_{th}$ ring zone serves as the outer edge for the $(j+1)_{th}$ zone. Like the target coordinate, the transition nodes lack features and labels (PM2.5 concentration). They serve as the intermediary states and springboards in the process of estimating $\hat{\mathbf{y}}^{tar}$. By increasing the value of $m$, the Ring Estimation block facilitates the inference in an approximately continuous manner. 

\subsection{Neighbor Aggregation}\label{st weight}
It is advisable to assign varying weights $\mathbf{W} = \left[ w1,\hdots,w_K \right]$  to the estimations of the target nodes based on the different spatio-temporal scenarios in which they are situated. To this end, we present the Neighbor Aggregation module, which takes into account the coding of the coordinates of the neighbors and the target coordinate and employs end-to-end learning to compute the estimation weights $w_i$ of each neighbor on the target node. 
To obtain $\mathbf{W}$, we first multiplied the output by $W_N \in \mathbb{R}^{10\times1}$ to transform it into a Logit score. Then, we applied the $Softmax$ operation to ensure that the weights sum up to one. In this end, the formulation of the Neighbor Aggregation can be written as
\begin{equation}
    \mathbf{W} = Softmax\left( W_N \cdot Decoder(\mathbf{P}^{src}, \mathbf{P}^{tar}) \right)
\end{equation}


\section{Experiments}
In this section, we delve into our experimental methodology aimed at evaluating the performance and validating the efficacy of the STFNN. 
Specifically, our experiments are designed to explore the following research questions, elucidating key aspects of our approach and its applicability in real-world scenarios:
\begin{itemize}
    \item \textbf{RQ1}: How does STFNN's approach, focusing on inferring concentration gradients for indirect concentration value inference, outperform traditional methods in terms of accuracy and effectiveness?
    \item \textbf{RQ2}: What specific contributions do the individual components of STFNN make to its effectiveness in inferring air pollutant concentrations?
    \item \textbf{RQ3}: How do variations in each hyperparameter impact the overall performance of STFNN?
    \item \textbf{RQ4}: Do the three-dimensional hidden states learned by the model accurately represents the gradient of the spatio-temporal field?
    \item \textbf{RQ5}: Can our model demonstrate proficient performance in inferring concentrations of various air pollutants, including \chemfig{NO_2}?
\end{itemize}
\begin{table*}[!ht]
    \centering
    \footnotesize
    \renewcommand{\arraystretch}{1.3}
    \setlength{\tabcolsep}{2.5pt}
    
    \resizebox{\textwidth}{!}{
        \begin{tabular}{c|c|l|cccc|cccc|cccc}
            
            \hline
            \multirow{2}{*}{Model} & \multirow{2}{*}{Year} & \multirow{2}{*}{\#Param(M)} & \multicolumn{4}{c|}{Mask Ratio = 25\%} & \multicolumn{4}{c|}{Mask Ratio = 50\%} & \multicolumn{4}{c}{Mask Ratio = 75\%} \\
            \cline{4-15}
            & & & MAE & $\boldsymbol{\Delta}$ & RMSE & MAPE & MAE & $\boldsymbol{\Delta}$ & RMSE & MAPE & MAE & $\boldsymbol{\Delta}$ & RMSE & MAPE \\
            \hline
            KNN & 1967 & \multicolumn{1}{c|}{\multirow{2}{*}{\begin{tabular}[c]{@{}c@{}}-\\- \\ \\ \end{tabular}}} & 30.50 & +146.0\% & 65.40 & 1.36 & 30.25 & +145.5\% & 72.23 & 0.71 & 34.07 & +194.0\% & 74.55 & 0.64 \\
            RF & 2001 & \multicolumn{1}{c|}{} & 29.22 & +135.6\% & 68.95 & 0.76 & 29.71 & +141.2\% & 71.61 & 0.75 & 29.82 & +157.3\% & 70.99 & 0.74 \\
            \hline
            MCAM & 2021 & \multicolumn{1}{c|}{0.408} & 23.94 & +93.1\% & 36.25 & 0.95 & 25.01 & +103.0\% & 37.94 & 0.92 & 25.19 & +117.3\% & 37.82 & 1.04 \\
            \hline
            SGNP & 2019 & \multicolumn{1}{c|}{\multirow{2}{*}{\begin{tabular}[c]{@{}c@{}}0.114\\0.108 \\ \\ \end{tabular}}} & 23.60 & +90.3\% & 37.58 & 0.83 & 24.06 & +95.3\% & 37.08 & 0.93 & 21.68 & +87.1\% & 33.68 & 0.84 \\
            STGNP & 2022 & & 23.21 & +87.2\% & 38.13 & 0.62 & 21.95 & +78.2\% & 37.13 & 0.67 & 19.58 & +68.9\% & 31.95 & 0.69 \\
            \hline
            VAE & 2013 & \multicolumn{1}{c|}{\multirow{2}{*}{\begin{tabular}[c]{@{}c@{}}0.011\\0.073 \\0.073 \\ \end{tabular}}} & 28.49 & +129.8\% & 67.11 & 0.94 & 28.92 & +134.7\% & 69.67 & 0.94 & 29.00 & +150.2\% & 69.11 & 0.93 \\
            GAE & 2016 & & 12.63 & +1.9\% & 23.80 & 0.46 & 12.78 & +3.7\% & 24.11 & 0.46 & 12.57 & +8.5\% & 23.73 & 0.46 \\
            GraphMAE & 2022 & & 12.40 & - & 23.20 & 0.46 & 12.32 & - & 23.11 & 0.46 & 11.59 & - & 21.51 & 0.43 \\
            \hline
            \rowcolor{gray!30} STFNN & - & \multicolumn{1}{c|}{0.208} & \textbf{11.14} & \textbf{-10.2\%} & \textbf{19.75} & \textbf{0.39} & \textbf{11.32} & \textbf{-8.1\%} & \textbf{19.91} & \textbf{0.42} & \textbf{11.27} & \textbf{-2.8\%} & \textbf{19.86} & \textbf{0.41} \\
            \hline
        \end{tabular}
    }
    \caption{Model comparison on the nationwide dataset. The parameter count, denoted as \#Param, is in the order of million (M). The symbol $\Delta$ represents the reduction in MAE compared to GraphMAE. The mask ratio represents the proportion of unobserved nodes to all nodes.}\label{main table}
\end{table*}
\subsection{Experimental Settings}
\subsubsection{Datasets}
The study obtained a nationwide air quality dataset~\cite{liang2023airformer} from January 1st, 2018, to December 31st, 2018. This dataset includes air quality and meteorological data. The input data can be divided into two classes: continuous and categorical data. Continuous data includes critical parameters such as air pollutant concentrations (e.g., PM2.5, CO), temperature, wind speed, and others. Categorical data encompasses weather, wind direction, and time. In the event of an unanticipated occurrence, such as a power outage, some data may be unavailable. 


\subsection{Baselines for Comparison}
We compare our STFNN with the following baselines that belong to the following four categories:
\begin{itemize}[leftmargin=*]
\item \textbf{Statistical models}: \textbf{KNN} \cite{guo2003knn} utilizes non-parametric, instance-based learning, inferring air quality by considering data from the nearest neighbors. \textbf{Random Forest (RF)} \cite{fawagreh2014random} aggregates interpolation from diverse decision trees, each trained on different dataset subsets, providing robust results.
\item \textbf{Neural Network based models}: \textbf{MCAM} \cite{han2021fine} introduces multi-channel attention blocks capturing static and dynamic correlations. 
\item \textbf{Neural Processes based models}: \textbf{SGNP}, a modification of Sequential Neural Processes (SNP) \cite{singh2019sequential}, incorporates a cross-set graph network before aggregation, enhancing air quality inference. \textbf{STGNP} \cite{hu2023graph} employing a Bayesian graph aggregator for context aggregation considering uncertainties and graph structure.
\item \textbf{AutoEncoder based models}: \textbf{VAE} \cite{kingma2022autoencoding} applies variational inference to air quality inference, utilizing reconstruction for target node inference. \textbf{GAE} \cite{kipf2016variational} reconstructs node features within a graph structure, while \textbf{GraphMAE} \cite{hou2022graphmae} introduces a masking strategy for innovative node feature reconstruction.
\end{itemize}

\subsection{Hyperparameters \& Setting}
To mitigate their impact, instances exceeding a 50\% threshold of missing data at any given time were prudently omitted from our analysis. Our dataset was carefully partitioned into three segments: a 60\% training set, a 20\% validation set, and a 10\% test set. During training, in each epoch, we randomly select stations with ratio $\alpha$, mask their features and historical information, and let them act as the target node. We ignore all locations where PM2.5 (or \chemfig{NO_2} in the case of \textbf{RQ5}) is missing. 
The model is trained with an Adam optimizer, starting with a learning rate of 1E-3, reduced by half every 40 epochs during the 200 training epochs. The batch size for training is set to 32. The hidden dimension of MLP and all the Transformer-Decoder networks is fixed at 64. For Ring Estimation, the neighbor number is set to 6, incorporating the past 6 timesteps, and the iteration step $m$ is defined as 16.

\subsection{Model Comparison (RQ1)}
In addressing RQ1, we conduct a meticulous comparative analysis among models based on the evaluation metrics of MAE, RMSE, and MAPE. The empirical outcomes derived from this analysis are systematically presented across the expansive spectrum of the nationwide air quality dataset, meticulously documented within Table \ref{main table}.

The results indicate that STFNN consistently demonstrates enhanced efficacy across various evaluation metrics, surpassing existing baseline models. Table \ref{main table} shows that our approach reduces MAE under three different mask ratios (25\%, 50\%, and 75\%) in comparison to GraphMAE, establishing a new State-of-the-Art (SOTA) in nationwide PM2.5 concentration inference in the Chinese Mainland. 
In our view, there are three main reasons for this. First, the gradient field is a better representation of reality. Second, the spatial and temporal modules of STFNN capture both types of information together, avoiding bias or information loss. Finally, our Pyramidal Inference framework captures global and local spatio-temporal properties, which helps us model the pollutant concentration field more accurately.

\begin{figure}[!b]
  \centering
  \includegraphics[width=\linewidth]{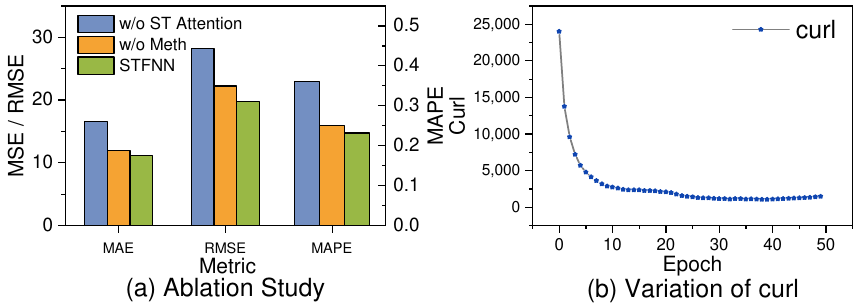}
  \caption{(a) ablation study (b) the variation of curl}
  \label{ab curl fig}
\end{figure}
\subsection{Ablation Study (RQ2)}
To assess the contributions of individual components to the performance of our model and address RQ2, we conducted ablation studies. The findings from these studies are presented in Figure \ref{ab curl fig} (a). 

\noindent \textbf{Effects of meteorological features.} To analyze the impact of meteorological features on the accuracy of the final model, we removed them from the raw data. Therefore, the gradient was obtained solely from the spatio-temporal coordinates of neighboring stations fed into the gradient encoder. The figure demonstrates that removing the meteorological features resulted in some improvement in the model's mean absolute error (MAE), which still outperformed all baseline models. 

\noindent \textbf{Effects of Neighbor Aggregation.} To investigate the impact of a dynamic and learnable implicit graph structure on the model, we substituted the model's Neighbor Aggregation module with IDW and SES, a non-parametric approach inspired by Zheng et al \cite{zheng2013u}. This approach employs implicit graph relations that are static. The results depicted in Figure \ref{ab curl fig} (a) demonstrate that utilizing the Neighbor Aggregation module results in a significant decrease in MAE.

\subsection{Hyperparameters Study (RQ3)}
In this section, we comprehensively explore the effects of various hyperparameters on the model's performance, thereby addressing RQ3.

\begin{figure}[!t]
    \centering
    \includegraphics[width=\linewidth]{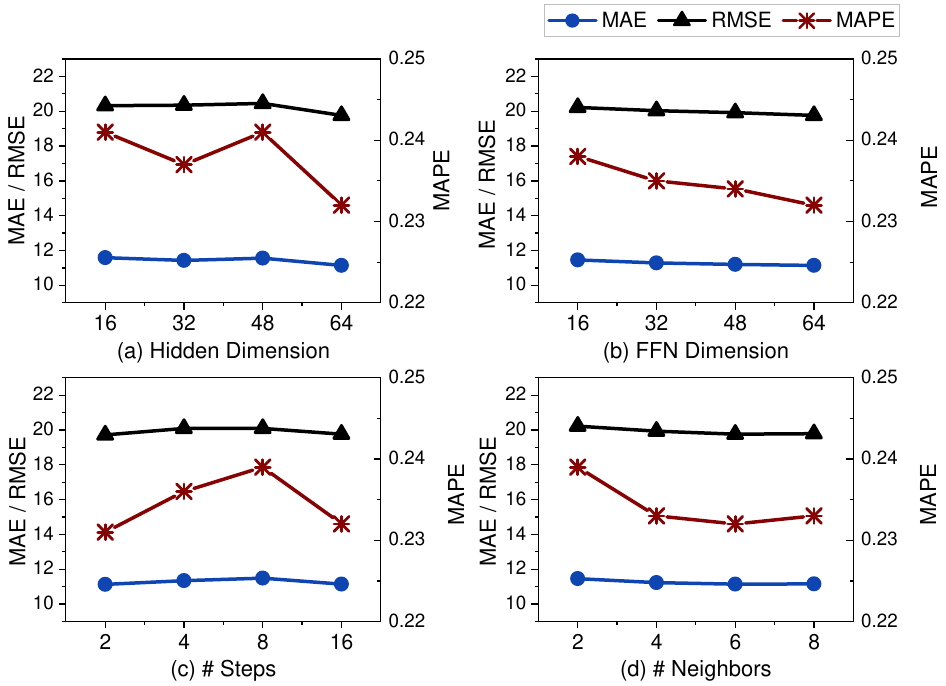}
    \caption{Hyperparameter Study}
    \label{hp study}
\end{figure}

\noindent \textbf{Effects of Hidden \& FFD Dimension.} We adjusted the hidden layer dimension of the Spatio-Temporal Encoding module and the forward propagation of the Transformer-Decoder structure used by the Ring Estimation and Neighbor Aggregation modules from 16 to 64. The results in Figure \ref{hp study} (a-b) show that adjusting the hidden layer dimension has little effect on the absolute values of MAE and RMSE, but it significantly decreases MAPE. 

\noindent \textbf{Effects of Step Size.} We vary the value of the accumulation step size of the Ring Estimation in $\left\{2,4,8,16\right\}$. The result is shown in Figure \ref{hp study} (c). It has been observed that as the step size increases, the model's performance initially declines before improving. Additionally, when comparing 2 steps to 16 steps, we note that the model's training time per round is approximately 50\% longer for 16 steps. 


\noindent \textbf{Effects of Neighbors Number.} We vary the number of neighbors from 2 to 8, and the result is shown in Figure \ref{hp study} (d). We observe that the performance of our model improved as the number of neighbors increased. Notably, our proposed STFNN exhibited excellent performance even with a small number of neighbors. Due to its ability to learn global spatio-temporal patterns, the model can use global information for inference even in scenarios where there are only a few neighbors present. 

\subsection{Interpretability (RQ4)} \label{curl experiment}
To confirm that the network's learned vector is the gradient of the spatio-temporal field, we calculate the curl variation with the number of training epochs. We use Yang et al.'s method \cite{yang2023neural} to quantify the curl and present the experimental results in Figure \ref{ab curl fig} (b). It is evident that the curl of the vector field obtained by the network decreases as the training progresses, indicating successful learning of the gradient field.
\begin{table}[!b]
    \centering
    \footnotesize
    \renewcommand{\arraystretch}{1.4}
    \setlength{\tabcolsep}{1.5pt}
    
    \resizebox{\linewidth}{!}{
        \begin{tabular}{c|cc|cc|cc}
            
            \hline
            \multirow{2}{*}{Model} & \multicolumn{2}{c|}{Mask Ratio = 25\%} & \multicolumn{2}{c|}{Mask Ratio = 50\%} & \multicolumn{2}{c}{Mask Ratio = 75\%} \\
            \cline{2-7}
            & MAE & RMSE & MAE & RMSE & MAE & RMSE \\
            \hline
            KNN & 18.10 & 62.51 & 18.47 & 64.22 & 20.18 & 62.86 \\
            RF & 16.90 & 61.25 & 17.60 & 64.91 & 17.36 & 63.70 \\
            \hline
            MCAM & 18.25 & 27.80 & 17.75 & 27.42 & 21.17 & 29.41 \\
            \hline
            SGNP & 17.66 & 25.43 & 19.17 & 26.36 & 16.57 & 24.11 \\
            STGNP & 16.43 & 27.85 & 15.62 & 26.06 & 15.70 & 26.23 \\
            \hline
            VAE & 29.85 & 112.81 & 31.43 & 119.59 & 30.82 & 117.33 \\
            GAE & 12.80 & 30.16 & 12.77 & 30.16 & 12.77 & 30.00 \\
            GraphMAE & 12.76 & 30.25 & 12.60 & 29.53 & 12.48 & 29.30 \\
            \hline
            \rowcolor{gray!30} STFNN & \textbf{11.34} & \textbf{23.65} & \textbf{11.52} & \textbf{24.93} & \textbf{11.97} & \textbf{25.81} \\
            \hline
        \end{tabular}
    }
    \caption{Experiment result on NO2}
    \label{no2 table}
\end{table}

\subsection{Generalizability (RQ5)}
Our model not only excels in inferring PM2.5 but also establishes a new benchmark, achieving the SOTA in inferring the concentration of \chemfig{NO_2}, as shown in Table \ref{no2 table}. This noteworthy outcome underscores the versatility of our model across distinct air quality parameters. In comparison to the baseline, our model demonstrates a significant advantage, showcasing its capability to handle diverse pollutants effectively and outperforming established methods in inference for \chemfig{NO_2}.

\section{Related Works}
Traditional methods~\cite{hasenfratz2014pushing,jumaah2019air} rely on linear spatial assumptions. However, these models only consider simple spatial relationships and do not adapt to complex changes in air quality. 
In recent times, there's been a growing interest in studying Spatio-Temporal Graph to understand the intricate relationship that involves both spatial and temporal for air quality inference. STGNNs~\cite{jiang2021dl,salim2015urban,wang2020deep,sun2020predicting,wang2021spatio}, which integrate the strengths of GNNs, have emerged as the leading approach for uncovering intricate relationships in STG data. Some follow-ups~\cite{li2017diffusion,yu2017spatio,geng2019spatiotemporal} introduce temporal components such as Recurrent Neural Networks (RNN)~\cite{graves2013generating} and Temporal Convolutional Networks (TCN)~\cite{bai2018empirical} to better address the spatio-temporal dependencies. However, The limitation of STGNNs lies in their lack of consideration for contiguity and Euclidean spatial structures.

\section{Conclusion}
In this work, we introduced a novel perspective for air quality inference, framing it as a problem of reconstructing Spatio-Temporal Fields (STFs) to better capture the continuous and unified nature of air quality data. Our proposed Spatio-Temporal Field Neural Network (STFNN) breaks away from the limitations of Spatio-Temporal Graph Neural Networks (STGNNs) by focusing on implicit representations of gradients, offering a more faithful representation of the dynamic evolution of air quality phenomena.  

\section*{Acknowledgements}
This work is supported by a grant from State Key Laboratory of Resources and Environmental Information System. This study is also funded by the Guangzhou-HKUST(GZ) Joint Funding Program
(No. 2024A03J0620). 

\bibliographystyle{named}
\bibliography{main}


\end{document}